\documentclass[letterpaper, 10 pt, conference]{ieeeconf}  

\IEEEoverridecommandlockouts                              

\usepackage{amsmath,amssymb,amsfonts}

\usepackage{amsthm}
\usepackage{subcaption}
\usepackage[english]{babel}
\usepackage{algorithmic}
\usepackage{algorithm}
\usepackage{graphicx}
\usepackage{textcomp}
\usepackage{xcolor}
\def\BibTeX{{\rm B\kern-.05em{\sc i\kern-.025em b}\kern-.08em
    T\kern-.1667em\lower.7ex\hbox{E}\kern-.125emX}}

\usepackage{censor}
\newif\ifanonymous
\anonymousfalse   

\usepackage[
    backend=biber,
    style=ieee,          
    citestyle=ieee-comp, 
    sorting=none         
]{biblatex}

\usepackage{subfiles}
\usepackage{multirow}

\usepackage[hidelinks]{hyperref}
\hypersetup{
  colorlinks=true,
  linkcolor=black,   
  citecolor=black,   
  urlcolor=magenta      
}
\usepackage{balance}
\usepackage[capitalize]{cleveref}

\usepackage[dvipsnames]{xcolor}
\newcommand{\blue}{\textcolor{blue}}
\newcommand{\red}{\textcolor{red}}
\newcommand{\cyan}{\textcolor{cyan}}
\newcommand{\orange}{\textcolor{orange}}
\newcommand{\green}{\textcolor{ForestGreen}}
\newcommand{\purple}{\textcolor{Purple}}

\title{\LARGE \bf
ELLIPSE: Evidential Learning for Robust Waypoints and Uncertainties
}

\ifanonymous
  \author{
    Anonymous Authors\\
    Anonymous Affiliations\\
    \thanks{Website: \url{https://EllipseStairTraversal.github.io}}
  }
\else
  \author{Zihao Dong$^{12}$, Chanyoung Chung$^{1*}$, Dong-Ki Kim$^{1*}$, Mukhtar Maulimov$^{1*}$, \\
    Xiangyun Meng$^1$, Harmish Khambhaita$^1$, Ali-akbar Agha-mohammadi$^1$, Amirreza Shaban$^1$
    \thanks{$^1$FieldAI. Work done while ZD intern at FieldAI.}%
    \thanks{$^2$Northeastern University, Boston, MA, USA.}
    \thanks{$^*$Equal Contribution, Alphabetical order by last name.}
    \thanks{Website: \url{https://EllipseStairTraversal.github.io}}
  }
\fi

\begin{document}

\maketitle
\thispagestyle{empty}
\pagestyle{empty}

\begin{abstract}

Robust waypoint prediction is crucial for mobile robots operating in open-world, safety-critical settings. 
While Imitation Learning (IL) methods have demonstrated great success in practice, they are susceptible to distribution shifts: the policy can become dangerously overconfident in unfamiliar states. 
In this paper, we present \textit{ELLIPSE}, a method building on multivariate deep evidential regression to output waypoints and multivariate Student-t predictive distributions in a single forward pass. 
To reduce covariate-shift-induced overconfidence under viewpoint and pose perturbations near expert trajectories, we introduce a lightweight domain augmentation procedure that synthesizes plausible viewpoint/pose variations without collecting additional demonstrations. 
To improve uncertainty reliability under environment/domain shift (e.g., unseen staircases), we apply a post-hoc isotonic recalibration on probability integral transform (PIT) values so that prediction sets remain plausible during deployment. 
We ground the discussion and experiments in staircase waypoint prediction, where obtaining robust waypoint and uncertainty is pivotal.
Extensive real world evaluations show that \textit{ELLIPSE} improves both task success rate and uncertainty coverage compared to baselines. 
\end{abstract}

\section{Introduction}

Trajectory or waypoint planning in open-world environments is a crucial capability for mobile robots, particularly in safety-critical domains such as construction, defense, and autonomous driving
\ifanonymous
    ~\cite{nahavandi2025comprehensive}.
\fi
Recent imitation learning (IL) approaches~\cite{sridhar2024nomad, yu2024trajectory, hu2024orbitgrasp, cheng2024navila} have demonstrated strong performance in predicting waypoint sequences from expert demonstrations.
However, learned waypoint predictors come with limited safety guarantees~\cite{koller2018learning, dong2024collision}, and can lead to catastrophic failures when deployed under distribution shift.

Uncertainty quantification (UQ) offers a principled mechanism to mitigate this risk by enabling a policy to recognize unreliable predictions and trigger conservative fallbacks (e.g., stopping and requesting expert assistance)~\cite{abdar2021review, he2025survey}.
In an ideal setting, higher uncertainty correlates with larger errors.
In robotics, however, limited demonstration data makes uncertainty estimates  vulnerable to covariate shift: the distribution of observations encountered during deployment can differ substantially from the training distribution, causing the model to remain overconfident when the prediction is wrong~\cite{ovadia2019can}.

\begin{figure}[t] 
    \centering
    \begin{subfigure}[b]{0.23\linewidth}
        \centering
        \includegraphics[width=0.95\columnwidth]{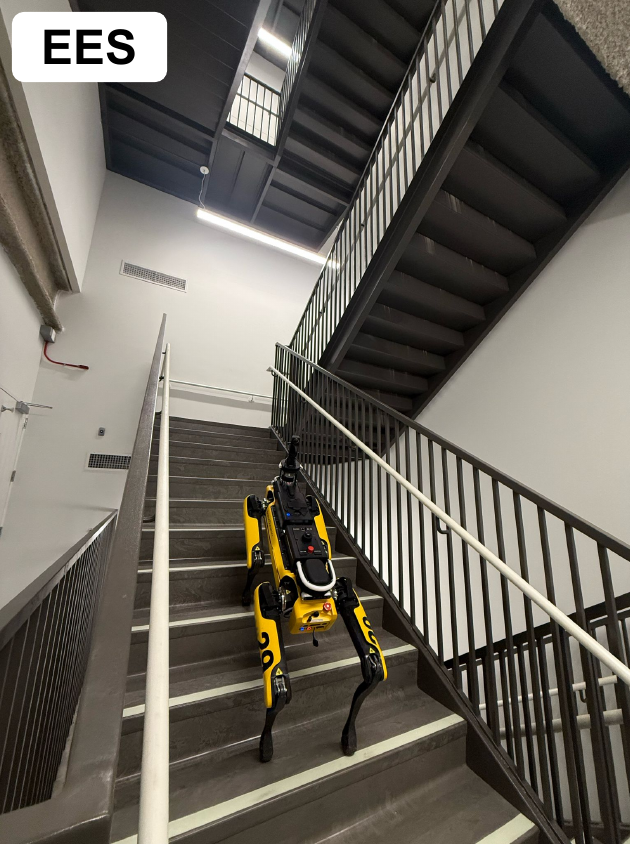}
        \vspace{-1.25em}
        \caption{EES}
        \label{fig:intro:env:ees}
    \end{subfigure}%
    \begin{subfigure}[b]{0.23\linewidth}
        \centering
        \includegraphics[width=0.95\columnwidth]{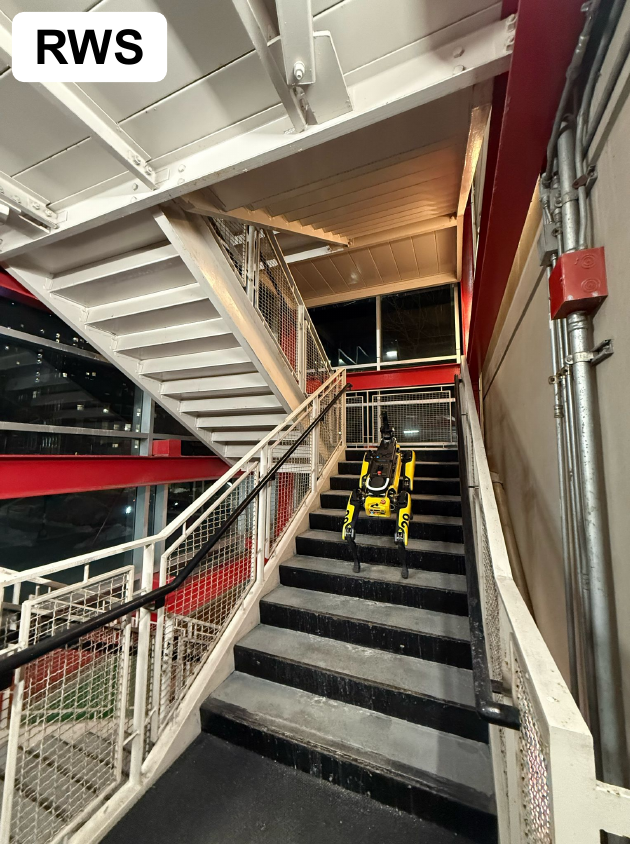}
        \vspace{-1.25em}
        \caption{RWS}
        \label{fig:intro:env:rws}
    \end{subfigure}%
    \begin{subfigure}[b]{0.23\linewidth}
        \centering
        \includegraphics[width=0.95\columnwidth]{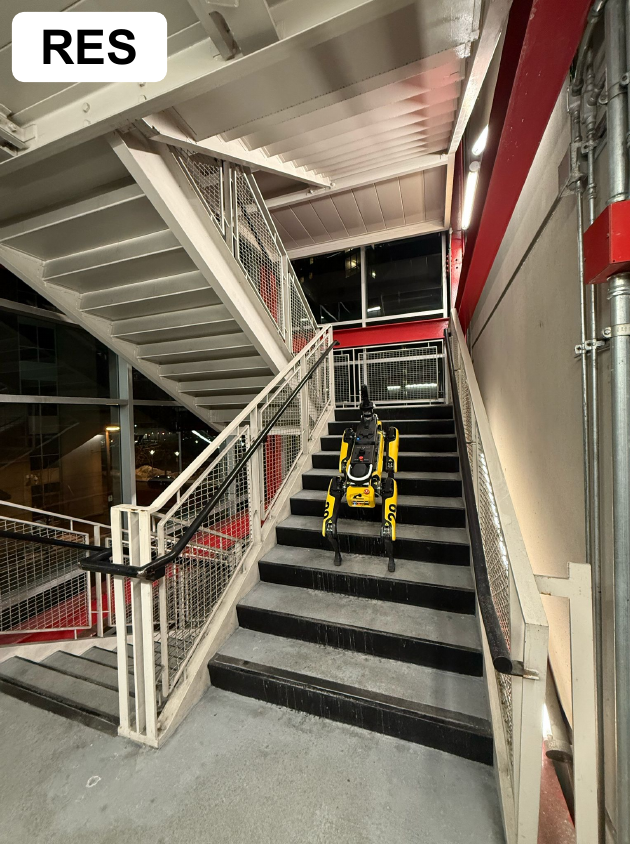}
        \vspace{-1.25em}
        \caption{RES}
        \label{fig:intro:env:res}
    \end{subfigure}%
    \begin{subfigure}[b]{0.23\linewidth}
        \centering
        \includegraphics[width=0.95\columnwidth]{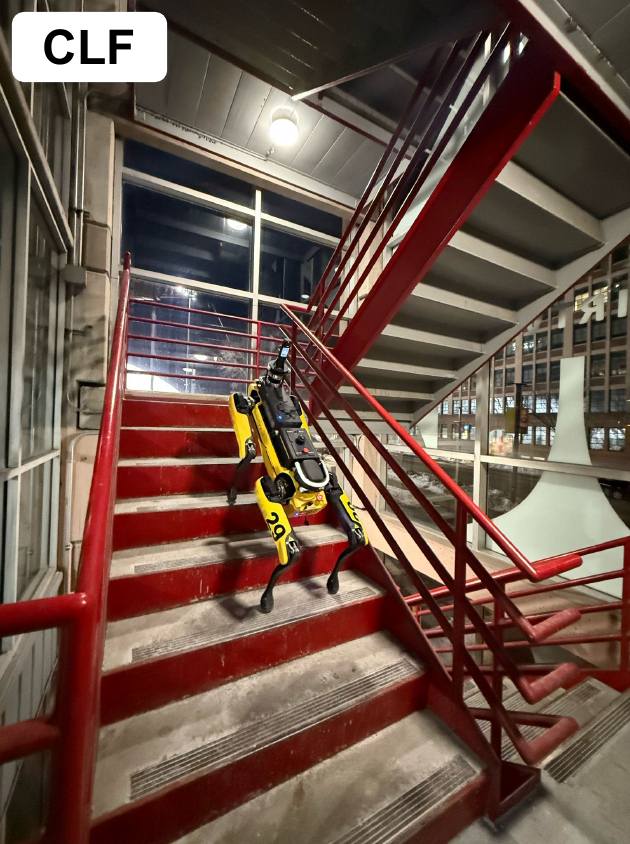}
        \vspace{-1.25em}
        \caption{CLF}
        \label{fig:intro:env:clf}
    \end{subfigure}
    \vspace{-0.5em}
    \caption{
        Deployment environments (\cref{sec:exp:setup}).
        Such environments are especially challenging due to limited sensor FOV, narrow passageways (landings), and invisible stair boundaries (hollow handrails and glass).
        The ability to navigate stairs is pivotal for robots to efficiently explore multi-floor structures.
    }
    \label{fig:intro:env}
    \vspace{-2em}
\end{figure} 

Stair navigation is a crucial capability for robots to safely explore multi-floor structures (e.g. construction sites), and it is a canonical scenario where accurate uncertainty estimation matters (\cref{fig:intro:env}).
First, staircase geometry---narrow passages, turns at landings, and elevation changes---restricts visibility and induces partial observability.
This motivates us to design a LiDAR-based method for its wider field of view.
Second, the margin for error is small: slight waypoint deviations can lead to severe consequences.
Last but not least, cascading error can easily drive the robot into viewpoints or poses off the (sparse) demonstration manifold, where the policy is wrong yet confident.
Beyond this learner-induced shift, deployment in novel staircases (e.g., different step geometry, materials, and sensing conditions) further induces environment/domain shift that degrades both waypoint and uncertainty reliability.

A popular approach to address covariate shift is to calibrate uncertainty online so that prediction sets maintain desirable coverage of the ground truth waypoints~\cite{tibshirani2019conformal, angelopoulos2021gentle, bastani2022practical}.
Such approaches rely on access to a conformity signal during deployment---for instance, the distance between the predicted and the ground truth waypoints---to update thresholds~\cite{ zhao2024conformalized}.
However, obtaining ground truth waypoints online typically requires human annotation, which is costly and error-prone, limiting the application of such methods in real deployments.

\begin{figure*}[t] 
    \centering
    \includegraphics[width=1.75\columnwidth]{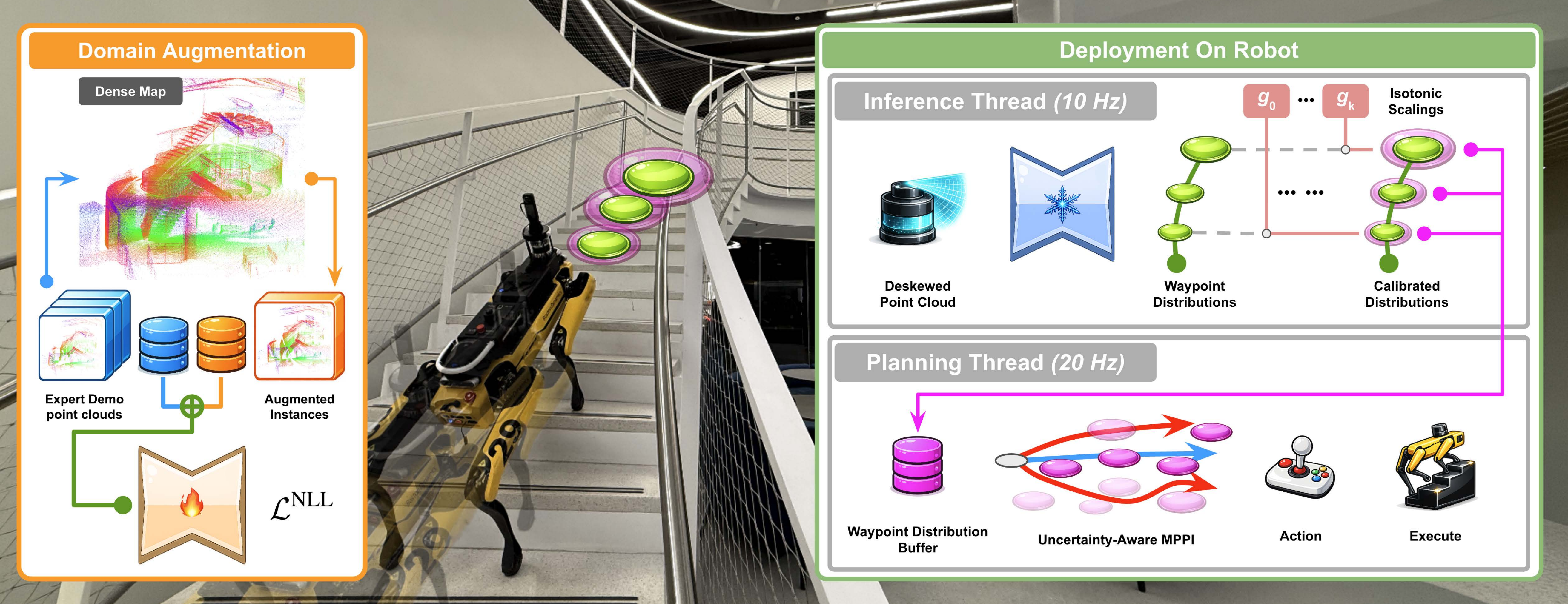}
    \caption{
        Overall pipeline of \textbf{ELLIPSE}.
        The network is trained offline with domain augmentation to improve robustness of the waypoints and uncertainties (\cref{sec:method:aug}).
        During inference, the predicted waypoint distributions (\cref{sec:method:der}) are recalibrated using scales obtained from Isotonic Regression (\cref{sec:method:isotonic}).
        A Mahalanobis-distance-based uncertainty-aware MPPI planner runs on a separate thread and tracks a pool of waypoints (at high frequency) while relaxing constraints on uncertain waypoints (\cref{sec:method:mppi}). 
    }
    \label{fig:pipeline}
    \vspace{-2em}
\end{figure*}

This motivates methods that produce reliable uncertainty without requiring online labels, while remaining lightweight enough for real-time deployment.
In this paper, we build on deep evidential regression (DER)~\cite{amini2020deep, meinert2021multivariate} to predict both waypoints and predictive distributions in a single forward pass. 
We believe it is more preferable in the problem setting than methods such as deep ensembles, which require multiple forward passes and thus can induce high inference latency~\cite{abdar2021review}.
To improve uncertainty reliability under viewpoint/pose shifts near the expert manifold, we introduce an effective domain augmentation procedure that synthesizes additional observations and corrective actions around expert trajectories, inspired by domain augmentation for lidar semantic segmentation~\cite{ryu2023instant} and demonstration synthesis for imitation learning~\cite{mandlekar2023mimicgen, zhou2023nerf, oh2025self}.
To address environment/domain shift (e.g., unseen staircases), we further recalibrate the predictive distribution using isotonic regression on probability integral transform (PIT) values~\cite{bramlage2023plausible}.
The calibrated distributions are integrated with an uncertainty-aware MPPI planner to generate plans that stay close to confident waypoints.
To summarize, we present \textbf{E}videntia\textbf{L} \textbf{L}earning for \textbf{I}nformative \textbf{P}robablistic Waypoint \textbf{SE}quences (\textbf{ELLIPSE}), with the following contributions:
\begin{itemize}
    \item An uncertainty-aware waypoint predictor based on multivariate deep evidential regression~\cite{meinert2021multivariate} that outputs waypoints and uncertainty in a single forward pass.
    \item A lightweight domain augmentation strategy that enlarges the support of the training distribution, improving robustness of waypoints and their uncertainties.
    \item A PIT-based isotonic recalibration procedure~\cite{kuleshov2018accurate, bramlage2023plausible} that improves coverage under environment/domain shift.
    \item Integration of the probabilistic waypoints with a Mahalanobis-distance-based MPPI planner.
    \item Extensive evaluations of \textbf{ELLIPSE} on staircase navigation, where it outperforms the baselines on both success rate and uncertainty coverage, and qualitative examples showcasing \textbf{ELLIPSE}'s practical benefits.
\end{itemize}

\section{Related Works}

\subsection{Uncertainty Quantification}
Uncertainty quantification (UQ) is critical for deploying modern deep learning methods in real-world, safety-critical settings~\cite{abdar2021review}. 
Common approaches include Monte Carlo (MC) Dropout~\cite{neal2012bayesian}, deep ensembles~\cite{zhou2025ensemble}, and evidential methods~\cite{meinert2021multivariate}.
While MC Dropout and ensembles are often effective, they require multiple forward passes, inducing latency prohibitive for real-time robotics.
In contrast, evidential deep learning produces the prediction, and associated aleatoric uncertainty (irreducible data noise) and epistemic uncertainty (reducible model uncertainty), in a single forward pass, making it appealing for on-robot deployment~\cite{meinert2021multivariate, cai2024evora, dong2025learning}.
Another widely used family of UQ methods is conformal prediction (CP)~\cite{angelopoulos2021gentle}. 
Online conformal variants are particularly attractive in robotics because they can adapt prediction set sizes under covariate shift while statistically guaranteeing marginal coverage~\cite{bastani2022practical, gibbs2024conformal}.
However, these online methods rely on access to a conformity signal during deployment, which for waypoint planning would require ground truth future waypoints—typically obtainable only through human annotation—making the approach costly and error-prone, limiting its practicality in our setting.

\subsection{Imitation Learning and Covariate Shift}
Imitation learning (IL), which trains policies from expert demonstrations, has achieved strong performance across a range of robotics tasks~\cite{sridhar2024nomad, yu2024trajectory, hu2024orbitgrasp, cheng2024navila}.
Despite this success, imitation policies are well known to be susceptible to \emph{covariate shift}, and the compounding error can lead to catastrophic consequences like collision and rollover~\cite{ross2011reduction, zhao2024conformalized}.
To mitigate this issue, we mainly identify two classes of methods.

\paragraph{Synthesizing Demonstrations}
Given a limited amount of expert demonstration, an appealing approach to train robust imitation learning policies is to synthesize new demonstrations to improve dataset support.
MimicGen shows that through re-purposing human demonstrations in new contexts, success rate for various table-top manipulation tasks can be significantly boosted~\cite{mandlekar2023mimicgen}.
SPARTN trains neural radiance fields from real world demonstrations to render observations from novel poses and generate corrective actions from those poses~\cite{zhou2023nerf}.
SART autonomously augments a single human demonstration using annotated safety region around trajectory keypoints~\cite{oh2025self}.
While these methods are mostly designed for camera and depth observations, we anticipate that the idea would transfer to LiDAR point clouds.

\paragraph{Interactive Imitation Learning}
Interactive IL (IIL) seeks to improve performance under the learner-induced state distribution by iteratively requesting for expert help during deployment and training on aggregated datasets~\cite{ross2011reduction, menda2019ensembledagger, zhao2024conformalized}.
In fact, uncertainty quantification (UQ) is widely adapted for robot-gated IIL, where the agent determines if it needs expert intervention and subsequently calls for help.
EnsembleDAgger~\cite{menda2019ensembledagger} leverages the ensemble disagreement (epistemic uncertainty), and ConformalDAgger~\cite{zhao2024conformalized} utilizes intermittent quantile tracking to calibrate prediction sets online. 
However, in the context of waypoint planning such methods are either too computationally expensive or impractical due to the lack of online ground truth.

\subsection{Placement of This Work}

Our method is complementary to the post-hoc methods like online CP and uncertainty-gated IIL~\cite{zhao2024conformalized, bastani2022practical, menda2019ensembledagger}: rather than adapting online, we improve the starting point of both the policy and its uncertainty estimates in an offline manner.
Our approach is also related in spirit to demonstration synthesis approaches (e.g., MimicGen, SPARTN)~\cite{mandlekar2023mimicgen, zhou2023nerf}.
Our focus, however, is LiDAR-based waypoint prediction with a specific emphasis on uncertainty reliability under covariate shift.
Although we evaluate ELLIPSE on staircase climbing with LiDAR, we hypothesize the overall recipe is applicable to other IL tasks and sensing modalities.

\section{Method}

In this section, we present ELLIPSE, a point-cloud-based model for predicting uncertainty-aware waypoint sequences from expert demonstrations.
The overall pipeline of ELLIPSE is shown in \cref{fig:pipeline}.
Our backbone is multivariate deep evidential regression ~\cite{meinert2021multivariate}, which produces both waypoints and multivariate Student-$t$ predictive distributions in a single forward pass (\cref{sec:method:der}).
To mitigate covariate-shift-induced overconfidence when the robot deviates from the demonstration manifold, we augment the training data by synthesizing plausible viewpoint and pose perturbations around each expert trajectory (\cref{sec:method:aug}).
Furthermore, we apply a lightweight post-hoc recalibration: we fit an isotonic regression mapping on probability integral transform (PIT) values so that the resulting prediction set sizes more faithfully adapt to the residual/error magnitudes during deployment (\cref{sec:method:isotonic}).
Finally, we integrate the predicted uncertainty into an MPPI planner, which leverages previously confident predictions to mitigate the impact of occasional poor predictions (\cref{sec:method:mppi}).

\subsection{Learning Waypoints with Deep Evidential Regression}
\label{sec:method:der}
In this work, we are interested in predicting a sequence of future waypoints and associated uncertainties using a point cloud input, and track the predicted waypoints with a motion planner.
Concretely, given a lidar point cloud\footnote{We omit timestamp subscript $t$ for cleanliness when possible.} $\mathbf{Q} \in \mathbb{R}^{m \times 3}$ where $m > 0$ denotes the number of points, we are interested in predicting a sequence of 2D waypoints in birds-eye-view (BEV), denoted $\mathbf{w}_{0:T} := \{\mathbf{w}_0, \mathbf{w}_1, \cdots, \mathbf{w}_T\}$ where $\mathbf{w}_i \in \mathbb{R}^n$.
We preprocess the waypoints to be equidistant with a stride of $d$ meters (in order to smooth out noise in the demonstration trajecotry), and each waypoint is assumed to be a sample drawn i.i.d. from a multivariate (bivariate in our case, i.e. $n=2$) Gaussian distribution with unknown mean $\boldsymbol{\mu}_i \in \mathbb{R}^n$ and unknown variance $\boldsymbol{\Sigma}_i \in \mathbb{R}^{n\times n}$.
The conjugate prior to this multivariate Gaussian likelihood is a Normal Inverse-Wishart (NIW) distribution:
\begin{equation} \label{eqn:method:niw}
    p(\boldsymbol{\mu}_i, \boldsymbol{\Sigma}_i) = NIW_i(\mathbf{m}_i).
\end{equation}
where $\mathbf{m}_i = \{\hat{\boldsymbol{\mu}}_i, \kappa_i, \boldsymbol{\Psi}_i, \nu_i\}$, $\hat{\boldsymbol{\mu}}_i \in \mathbb{R}^n, \kappa_i > 0, \nu_i > n+1, $ and $\boldsymbol{\Psi}_i \in \mathbb{R}^{n\times n}$ is symmetric positive definite.
The probability of observing the ground truth waypoint $\mathbf{w}_i$ admits a multivariate student-t distribution:
\begin{equation} \label{eqn:method:predictive}
    p(\mathbf{w}_i | \mathbf{m}_i) = t_{\nu_i-n+1}(\hat{\boldsymbol{\mu}}_i, \frac{1}{\nu_i-n+1} \frac{1+\kappa_i}{\kappa_i} \boldsymbol{\Psi}_i).
\end{equation}
Concretely, a neural network $\Gamma_{\boldsymbol{\theta}}$, where $\boldsymbol{\theta}$ denotes trainable parameters, outputs the NIW parameters for every waypoint $\{\hat{\boldsymbol{\mu}}_i, \kappa_i, \mathbf{L}_i, \nu_i\}$ where $\mathbf{L}_i \in \mathbb{R}^{n\times n}$ is a lower triangular matrix with positive diagonal elements and $\boldsymbol{\Psi}_i = \mathbf{L}_i\mathbf{L}_i^\intercal$, and is trained by minimizing the negative logarithm likelihood:
\begin{equation} \label{eqn:method:nll}
     \mathcal{L}^{\text{NLL}}(\boldsymbol{\theta}) = -\frac{1}{T} \sum_{i=0}^T \text{log } p(\mathbf{w}_i | \mathbf{m}_i).
\end{equation}
The predicted waypoints are thus the mean of the NIW $\mathbb{E}[\boldsymbol{\mu}_i] = \hat{\boldsymbol{\mu}}_i$, and the aleatoric and epistemic uncertainty can be computed following~\cite{meinert2021multivariate}.

Although multivariate deep evidential regression provides a simple and principled way to predict waypoints and uncertainty, prior work has shown that evidential uncertainty may behave more like a proxy for residual error than a faithful uncertainty estimate~\cite{meinert2023unreasonable}.
In particular, while the mean prediction $\hat{\boldsymbol{\mu}}_i$ can fail catastrophically with unfamiliar inputs, the learned uncertainty may not increase accordingly because it is optimized on in-distribution residuals. 
Consequently, the model can remain overconfident exactly in the regimes where its errors are largest, as we empirically demonstrate in \cref{sec:exp:coverage}.
This motivates our domain augmentation technique, whose goal is to expose the model to a broader set of states likely to be encountered during deployment.

\subsection{Domain Augmentation via Synthesizing Novel Viewpoints}
\label{sec:method:aug}

To synthesize new training instances that lie off the original human demonstration trajectory, we first generate LiDAR point clouds from novel viewpoints.
Specifically, for each demonstration frame $t$, we build a dense point cloud map by aggregating geometrically adjacent, deskewed LiDAR scans using the robot pose estimates from SLAM~\cite{chen2022direct, dong2025lidar}, and denote the resulting map by $\mathbf{Q}^{world}_t$.
To reduce contamination from dynamic objects, we apply a lightweight consistency-based heuristic filter to $\mathbf{Q}^{world}_t$.

We then sample a perturbed pose $\hat{\mathbf{T}}_t$ within a prescribed safety margin $\boldsymbol{\epsilon}$ around the current pose $\mathbf{T}_t$, and transform the aggregated map accordingly to obtain $\hat{\mathbf{Q}}^{world}_t = \hat{\mathbf{T}}_t^{-1} \circ \mathbf{T}_t \mathbf{Q}_t^{world}$.
To avoid computationally expensive point-cloud ray casting, we follow~\cite{ryu2023instant} and project $\hat{\mathbf{Q}}^{world}_t$ to a $V \times H$ range image, where $V$ and $H$ denote the target vertical and horizontal resolutions, respectively, and then back-project to obtain the synthesized LiDAR point cloud $\hat{\mathbf{Q}}_t$.
Consistent with our hardware setup, we additionally include a set of predefined planes (e.g., LiDAR roll cage and robot chassis) to simulate self-occlusion / self-hits during synthesis.

The corresponding ground-truth waypoint sequence $\hat{\mathbf{w}}_{t, 0:T} := \{\hat{\mathbf{w}}_{t, 0}, \hat{\mathbf{w}}_{t, 1}, \cdots, \hat{\mathbf{w}}_{t, T}\}$ is obtained by sampling equidistant poses along the future demonstration trajectory and projecting them to the robot-centric BEV frame.
Semantically, these augmented instances encourage the model to learn self-corrective behavior when it deviates from the nominal trajectory during deployment~\cite{zhou2023nerf}.
Examples of the dense map and generated point cloud are shown in \cref{fig:pipeline}.

Although our domain augmentation improves the robustness of waypoint predictions and uncertainties when the robot deviates into off-manifold states, it does not by itself guarantee reliable uncertainty at deployment.
Due to the limited and sparse nature of robotics demonstration data, imitation learning policies are prone to overfitting to the demonstration distribution~\cite{belkhale2023data}.
As a result, residuals (or prediction errors) are often larger during testing and deployment than during training.
However, the uncertainty estimates produced by deep evidential regression are learned primarily from in-distribution residual patterns observed during training, and may therefore be miscalibrated for the larger residual magnitudes encountered at deployment under covariate shift~\cite{bramlage2023plausible}.
This motivates a post-hoc recalibration step that scales the evidential predictive distribution (via PIT-based isotonic regression) so that prediction-set coverage better matches empirical residual magnitudes under shift.

\subsection{Calibrating Uncertainty with Isotonic Regression}
\label{sec:method:isotonic}
\begin{figure}[t] 
    \centering
    \begin{subfigure}[b]{0.45\linewidth}
        \centering
        \includegraphics[width=0.95\columnwidth]{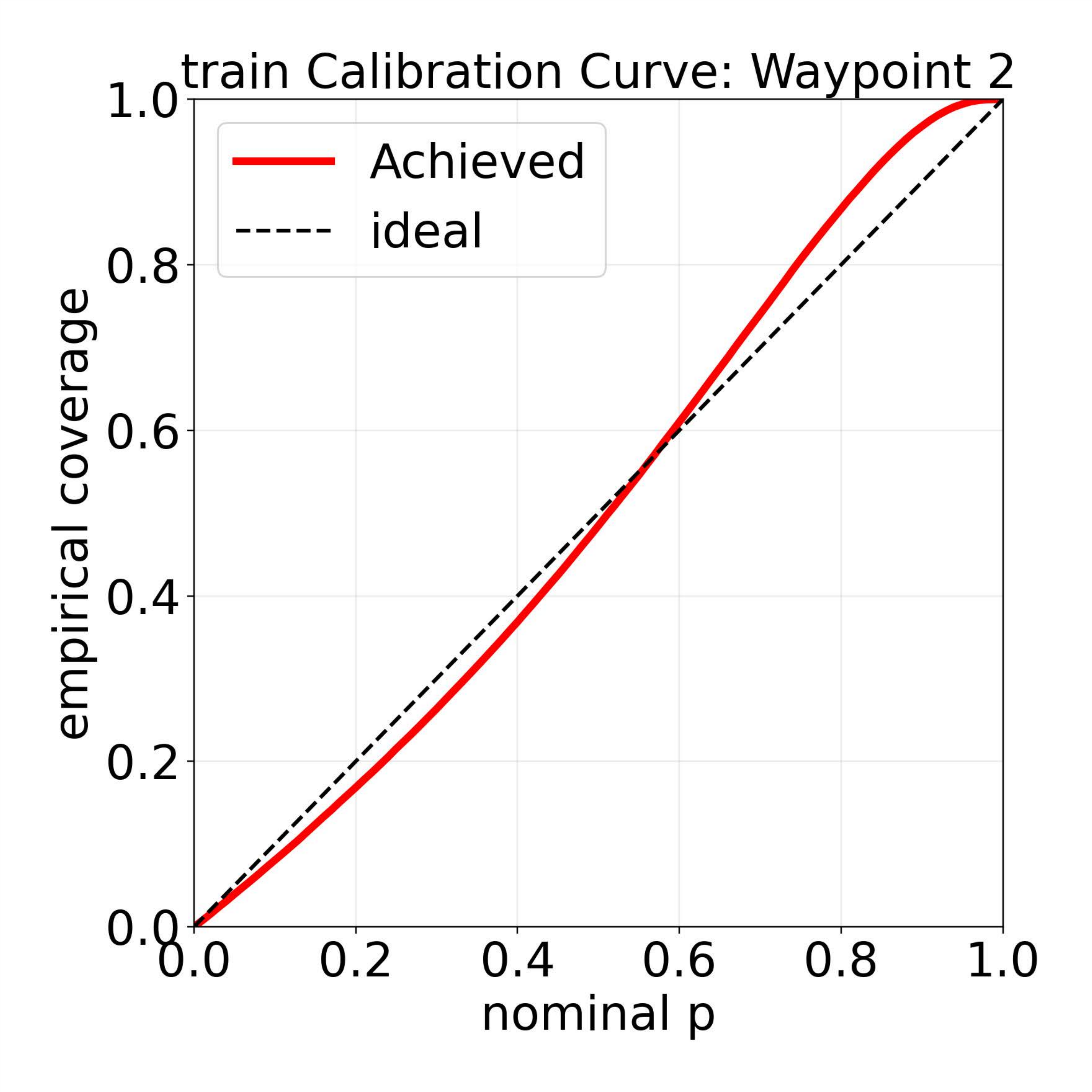}
        \vspace{-1em}
        \caption{Train calib. plot (waypoint 2)}
        \label{fig:method:calib_curve:train}
    \end{subfigure}%
    \begin{subfigure}[b]{0.45\linewidth}
        \centering
        \includegraphics[width=0.95\columnwidth]{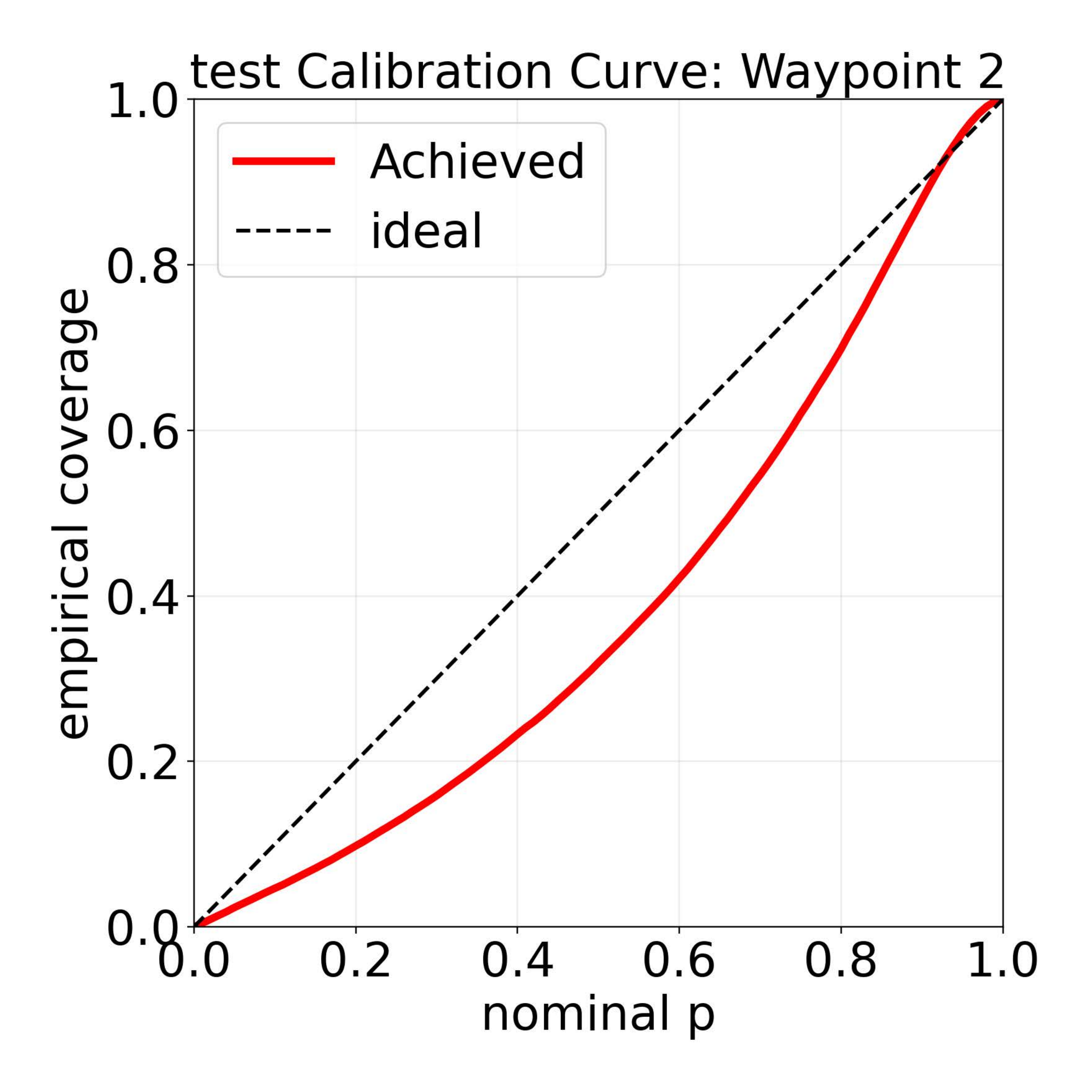}
        \vspace{-1em}
        \caption{Test calib. plot (waypoint 2)}
        \label{fig:method:calib_curve:test}
    \end{subfigure}
    \vspace{-0.5em}
    \caption{Uncertainty is calibrated on train yet overconfident on test.}
    \label{fig:method:calib_curve}
    \vspace{-1.8em}
\end{figure} 

In a regression setting, calibration means that a prediction set containing $p$\% of the predicted probability mass should contain the ground truth approximately $p$\% of the time~\cite{kuleshov2018accurate}. Equivalently, the calibration curve should lie close to the identity (diagonal) line.

In multivariate deep evidential regression, the posterior predictive for waypoint $\mathbf{w}_i$ is a multivariate Student-$t$ distribution with mean $\hat{\boldsymbol{\mu}}_i$, scale $\mathbf{S}_i = \frac{1}{\nu_i-n+1} \frac{1+\kappa_i}{\kappa_i} \boldsymbol{\Psi}_i$, and degree of freedom $\nu_i - n + 1$ (\cref{eqn:method:predictive}).
We calibrate this predictive distribution directly, rather than using a Gaussian approximation built from aleatoric/epistemic uncertainty terms as in~\cite{bramlage2023plausible}.
This choice keeps the calibration procedure consistent with the probabilistic model actually used by multivariate deep evidential regression at inference time.
To do so, we employ a radial probability integral transform (PIT), which maps each prediction--target pair to a scalar confidence value in $[0,1]$.

Specifically, for a ground-truth waypoint $\mathbf{w}_i$, we first compute the squared Mahalanobis radius under the predicted Student-t scale:
\begin{equation}
    r_i^2 = d_\mathrm{mah}(\mathbf{w}_i; \hat{\boldsymbol{\mu}}_i, \mathbf{S}_i) = (\mathbf{w}_i - \hat{\boldsymbol{\mu}}_i)^\top \mathbf{S}_i^{-1} (\mathbf{w}_i - \hat{\boldsymbol{\mu}}_i).
\end{equation}
For a multivariate Student-t predictive, $n^{-1}r^2_i$ admits an F-distribution, which allows us to compute a scalar PIT value
\begin{equation}
    u_i = F_i(n^{-1}r_i^2) \in [0,1],
\end{equation}
where $F_i$ is the predictive CDF for the $i$-th waypoint~\cite{roth2012multivariate}. 
If the predictive distribution is calibrated, then $u_i$ should be approximately uniformly distributed on $[0,1]$.
However, as shown in the calibration plot (\cref{fig:method:calib_curve}), while the training calibration curve (\cref{fig:method:calib_curve:train}) stays close to the diagonal, the testing curve (\cref{fig:method:calib_curve:test}) is mostly below the diagonal, signifying that the uncertainty during testing is overconfident.

Following ~\cite{kuleshov2018accurate, bramlage2023plausible}, we fit a monotonic recalibration mapping $g_i(\cdot): [0,1] \rightarrow [0, 1]$ using isotonic regression on a held-out calibration set $\{(u_{i, k}, \hat{p}_{i, k})\}_{k=1}^N$, where $\hat{p}_{i, k}$ denotes the percentage of data whose PIT is covered by $u_{i, k}$.
Intuitively, $g_i$ corrects systematic overconfidence (or underconfidence) in the raw evidential predictive distribution by applying a threshold-conditioned scaling, while preserving the ranking induced by the original predictive distributions\footnote{The resulting calibration curves should be close to the diagonal~\cite{kuleshov2018accurate}}.

At inference time, to construct a prediction set with nominal coverage level $p$, we invert the isotonic map to obtain the corresponding raw PIT threshold that empirically covers $p$\% of the samples, i.e. $\tilde{p} = g_i^{-1}(p)$, and form the Student-t ellipsoidal prediction set that covers $\tilde{p}$\% of the probability mass, yielding calibrated predictions sets whose empirical coverage more closely matches the desired coverage level\footnote{We make the assumption that the residual magnitude in calibration set is more closely aligned with that in the deployment set.}.

\subsection{Integrating Learned Uncertainty with Motion Planner}
\label{sec:method:mppi}
In order to execute the predicted waypoints, we approximate the robot's dynamics model with a unicycle model:
\begin{equation} \label{eqn:method:unicycle}
    \mathbf{x}_{t+1} = \text{unicycle}(\mathbf{x}_t, \mathbf{a}_t)
\end{equation}
where $\mathbf{x}_t = [x_t, y_t, \theta_t] \in \mathbf{X} \subseteq \mathbb{R}^3$ is the state vector containing the robot's BEV position (denoted $\mathbf{p}_t$) and orientation (yaw), and $\mathbf{a}_t = [v_t, \omega_t] \in \mathbf{A} \subseteq \mathbb{R}^2$ is the target linear and angular velocities.
An MPPI planner can be used to track a dense waypoint sequence $\mathbf{w}_{t, 0:H}$ (interpolated from the sparse waypoints $\mathbf{w}_{t, 0:T}$), where $H$ is the horizon~\cite{williams2016aggressive}.
However, this naive planner would:
\begin{itemize}
    \item Converge easily to local minima as the waypoints may not be feasible under the control limits,
    \item Fail to distinguish uncertain waypoints, and
    \item Discard past confident predictions that could help navigate occasional bad predictions.
\end{itemize}

As a result, we design a simple yet effective approach to include the predictive uncertainties into the MPPI framework to improve the robustness of the motion planner.
Our intuition is to leverage mahalanobis distance as opposed to traditional Euclidean distance cost in MPPI planners, which helps relax constraints near uncertain waypoints.

Intuitively, we absorb the isotonic scaling into the scale matrix $\mathbf{S}_{t,i}$ of the predictive multivariate student-t distribution, such that the $p$\% prediction set under the resulting distribution is equivalent to the $\tilde{p}$\% prediction set of the pre-isotonic distribution.
Let $\tilde{\mathbf{S}}_{t,i}$ denote the post-isotonic scale matrix, then it is computed as follow:
\begin{equation} \label{eqn:method:scale_S}
    \alpha_i = \frac{F^{-1}_i(\tilde{p})}{F^{-1}_i(p)}, \quad \quad
    \tilde{\mathbf{S}}_{t,i} = \alpha_i \, \mathbf{S}_{t,i}.
\end{equation}
Given the post-isotonic waypoint distributions \((\hat{\boldsymbol{\mu}}_{t,i}, \tilde{\mathbf{S}}_{t,i})\) and a rollout trajectory \(\mathbf{x}_{t,0:H}\), we first define a waypoint-tracking MPPI objective using the minimum Mahalanobis distance from the rollout to each predicted waypoint:
\begin{equation} \label{eqn:method:mppi_pre_norm}
    C(\mathbf{x}_{t,0:H})
    =
    \sum_{i=0}^{T}
    \min_{0 \le h \le H}
    d_{\mathrm{mah}}\!\left(\mathbf{p}_{t,h}; \hat{\boldsymbol{\mu}}_{t,i}, \tilde{\mathbf{S}}_{t,i}\right).
\end{equation}

In practice, we observe that many post-isotonic scale matrices \(\tilde{\mathbf{S}}_{t,i}\) have eigenvalues smaller than \(1\), which makes the corresponding Mahalanobis penalty more restrictive than Euclidean distance. 
As a result, even uncertain waypoints may remain overly punitive and fail to induce the desired relaxation in planning.
To better control this behavior, we introduce an expert-specified safety threshold \(\delta > 0\), shared across waypoints, defined in units of the semi-major axis length of the unit ellipse induced by \(\tilde{\mathbf{S}}_{t,i}\). Let
\begin{equation} \label{eqn:method:semi_major}
    \lambda_{t,i} = \mathrm{EigenVal}_{\max}(\tilde{\mathbf{S}}_{t,i}), 
    \qquad
    \ell_{t,i} = \sqrt{\lambda_{t,i}} .
\end{equation}
Here, \(\ell_{t,i}\) is the semi-major axis length of the unit ellipse. We then construct the scale matrix used in the MPPI cost, denoted \(\mathbf{S}^{\text{cost}}_{t,i}\), by applying no scaling when \(\ell_{t,i} \le \delta\), and an exponential relaxation when \(\ell_{t,i} > \delta\)\footnote{We do not discard such waypoints as they still carry useful information}:
\begin{equation} \label{eqn:method:normalize_s_piecewise}
    \mathbf{S}^{\text{cost}}_{t,i}
    =
    \begin{cases}
        \tilde{\mathbf{S}}_{t,i}, & \ell_{t,i} \le \delta,\\[4pt]
        \lambda_{t,i}^{-1}\left(\dfrac{\ell_{t,i}}{\delta}\right)^{\beta} \tilde{\mathbf{S}}_{t,i}, & \ell_{t,i} > \delta,
    \end{cases}
\end{equation}
where \(\beta \in \mathbb{R}_{+}\) controls the aggressiveness of the relaxation.

Under the relaxed branch (\(\ell_{t,i} > \delta\)), \cref{eqn:method:normalize_s_piecewise} rescales \(\tilde{\mathbf{S}}_{t,i}\) so that the largest eigenvalue of \(\mathbf{S}^{\text{cost}}_{t,i}\) becomes \(\left(\ell_{t,i}/\delta\right)^{\beta}\). Therefore, waypoint distributions with larger semi-major axes (relative to \(\delta\)) become progressively less punitive along their principal uncertainty direction, while waypoint distributions below the threshold remain unchanged.
Let $\tau \geq 1$ denote the number of historical predictions we consider, the MPPI planner then minimizes the following cost\footnote{We omit auxiliary costs like smoothness for brevity.}:
\begin{equation} \label{eqn:method:mppi}
    C(\mathbf{x}_{t,0:H})
    =
    \sum_{k=0}^{\tau}
    \sum_{i=0}^{T}
    \min_{0 \le h \le H}
    d_{\mathrm{mah}}\!\left(\mathbf{p}_{t,h}; \hat{\boldsymbol{\mu}}_{t-k,i}, \mathbf{S}^{\text{cost}}_{t-k,i}\right).
\end{equation}

\section{Experiments}

In this section, we evaluate ELLIPSE on stair traversal using a Boston Dynamics Spot with an Ouster OS0-128 LiDAR. All inference and planning are performed on a platform with 16 GB of unified memory, demonstrating that ELLIPSE is sufficiently lightweight for deployment on edge platforms. Through extensive quantitative and qualitative experiments we show the effectiveness of the proposed approach. 
In particular, this section addresses the following questions:

\begin{itemize}
\item Are the nominal waypoints reliable when directly tracked by MPPI, and does domain augmentation improve task success rate (\cref{sec:exp:sr})?
\item Do domain augmentation and isotonic recalibration improve empirical coverage at deployment (\cref{sec:exp:coverage})?
\item Does the proposed MPPI planner improve the robustness of the generated paths (\cref{sec:exp:planner})?
\end{itemize}

\subsection{Experiment Setup} \label{sec:exp:setup}
ELLIPSE takes as input a deskewed LiDAR point cloud from a SLAM pipeline~\cite{chen2022direct}. 
We process each point cloud by gravity-aligning it, clipping it to the axis-aligned box 
$[-10,-10,-4]\times[10,10,4]$, and randomly subsampling 20{,}000 points.
The resulting point cloud is encoded using a PointPillars backbone~\cite{lang2019pointpillars} with a pillar size of $0.16$\,m, whose output features are fed to a self-attention-based inpainting module followed by a ResNet encoder.
Finally, an MLP maps the learned feature representation to the waypoint predictions and their associated uncertainties.

For training, we collect demonstrations on 25 diverse staircases, of which 21 are used for training and 4 for testing.
We refer to the four test staircases (\cref{fig:intro:env}) as EES (7 floors, right-turning), RWS (10 floors, left-turning), RES (10 floors, left-turning), and CLF (7 floors, right-turning).
The ground-truth targets are $T=5$ equally spaced waypoints with stride $d=0.5$\,m.
Each training instance is augmented into 8 additional poses, and the safety margin is set to 
$\boldsymbol{\epsilon}=[\Delta_x,\Delta_y,\Delta_z,\Delta_{roll},\Delta_{pitch},\Delta_{yaw}]=[0,0.2,0.05,10,10,30]$,
with translational components measured in meters and rotational components in degrees.
We do not perturb along the robot $x$-axis, since this corresponds to perturbations along the demonstration trajectory and can introduce inconsistent point clouds that degrade training.
The model is trained for 50 epochs using a one-cycle scheduler with cosine annealing.
On our edge compute platform, the resulting model runs in real time on 10+\,Hz.

\subsection{Stair Climbing Success Rate on Unseen Stairs} \label{sec:exp:sr}
\begin{figure}[t] 
    \centering
    \begin{subfigure}[b]{0.45\linewidth}
        \centering
        \includegraphics[width=0.95\columnwidth]{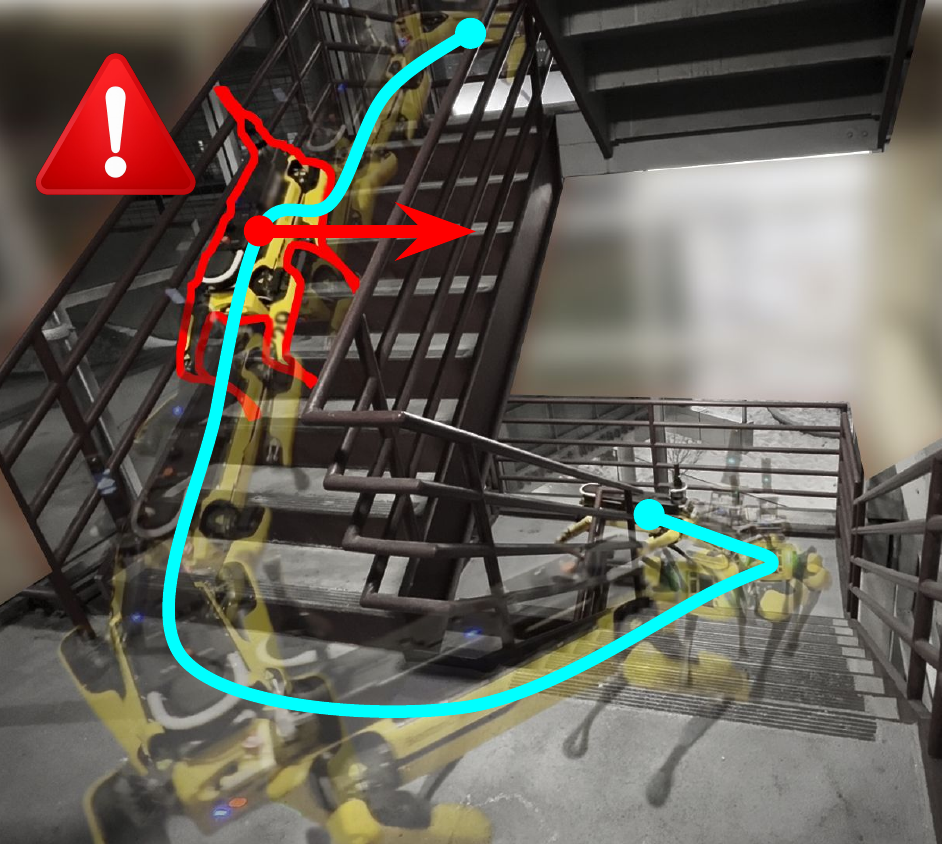}
        \vspace{-0.5em}
        \caption{w/o Aug. Path Following.}
        \label{fig:exp:overlay:no_aug_track}
    \end{subfigure}%
    \begin{subfigure}[b]{0.45\linewidth}
        \centering
        \includegraphics[width=0.95\columnwidth]{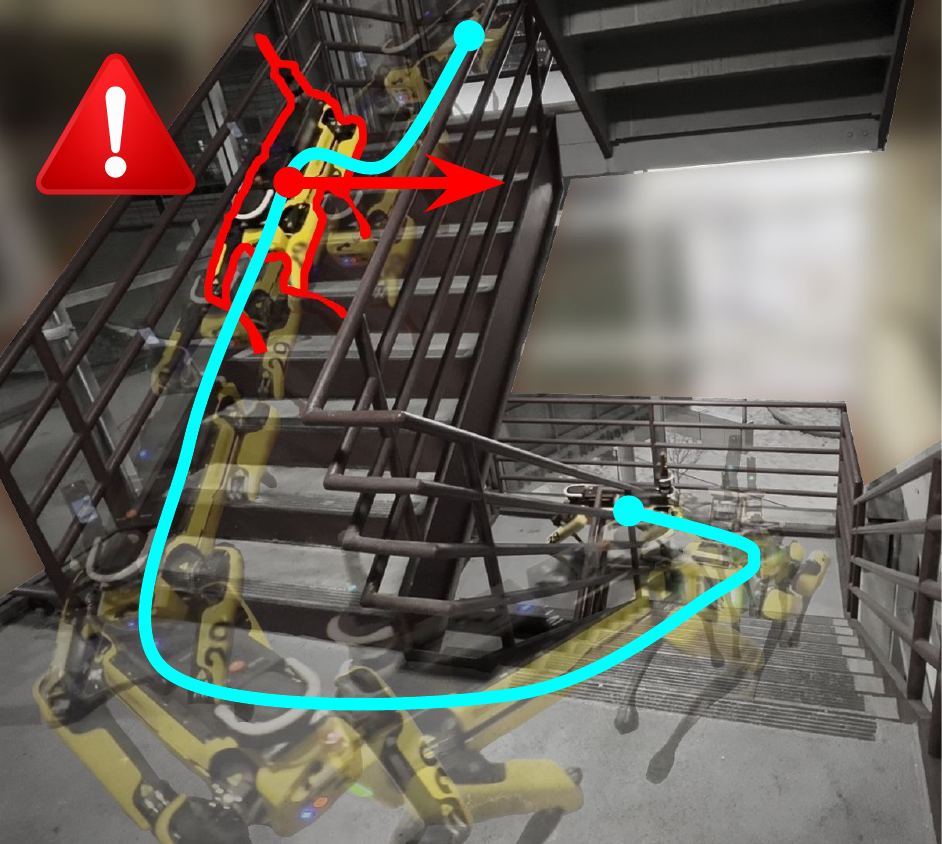}
        \vspace{-0.5em}
        \caption{w/o Aug. \cref{eqn:method:mppi} and $\tau=5$.}
        \label{fig:exp:overlay:no_aug_track_hist5}
    \end{subfigure}
    \begin{subfigure}[b]{0.45\linewidth}
        \centering
        \includegraphics[width=0.95\columnwidth]{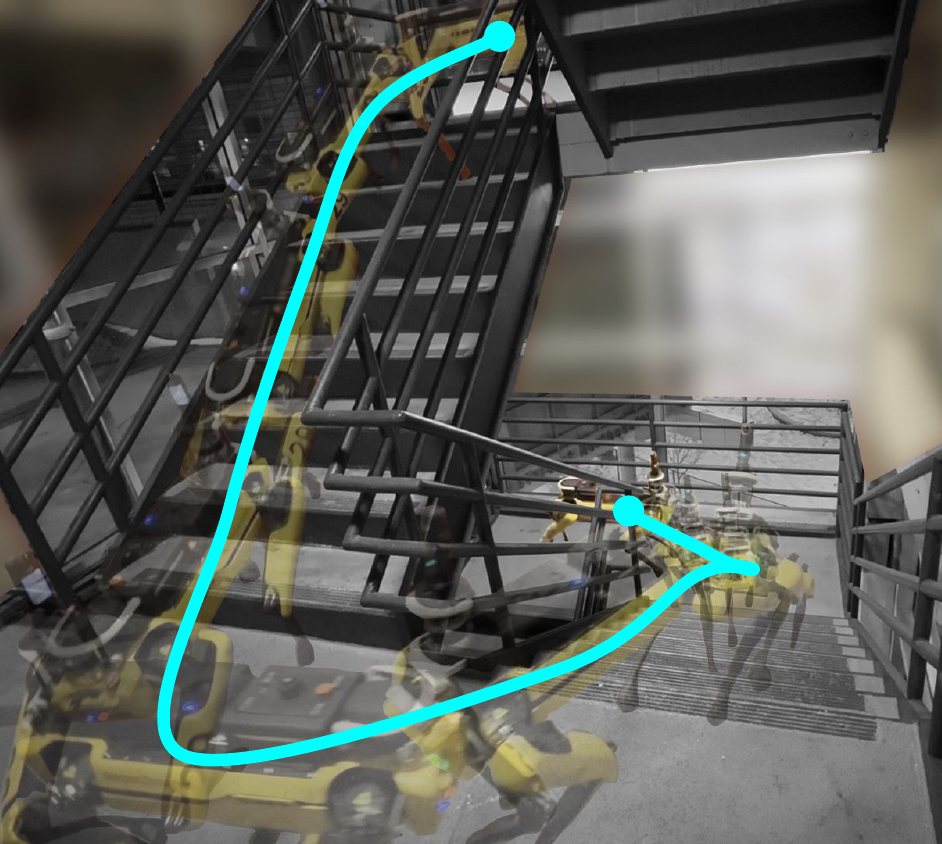}
        \vspace{-0.5em}
        \caption{w Aug. Path Following.}
        \label{fig:exp:overlay:aug_track}
    \end{subfigure}%
    \begin{subfigure}[b]{0.45\linewidth}
        \centering
        \includegraphics[width=0.95\columnwidth]{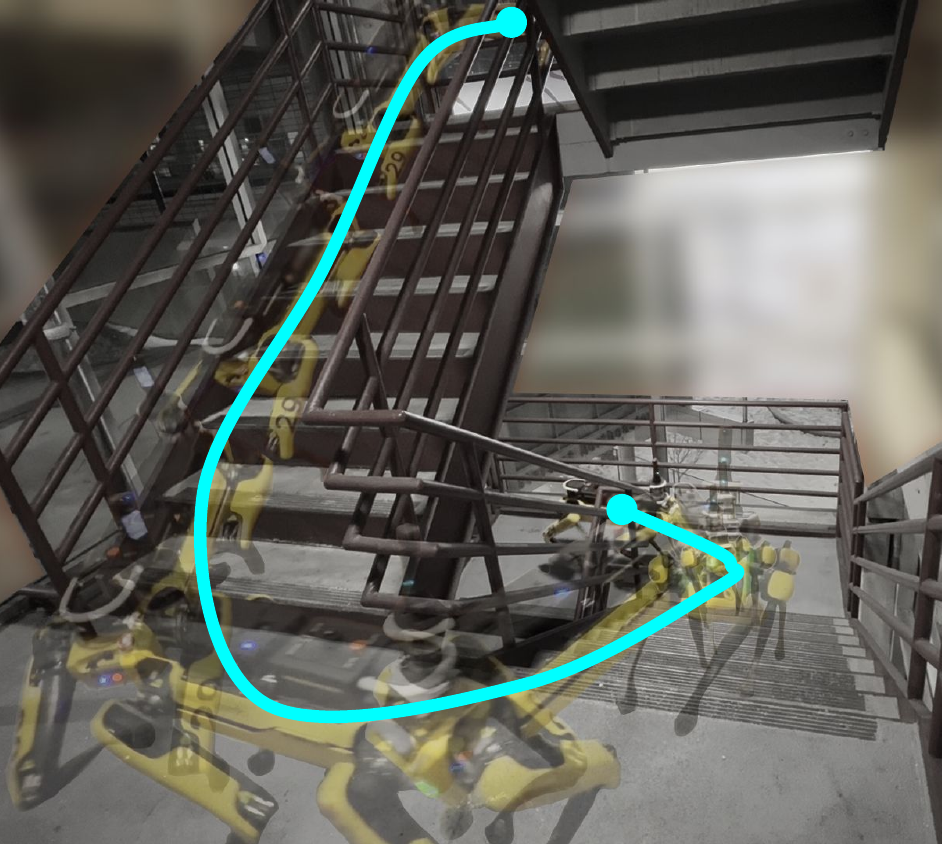}
        \vspace{-0.5em}
        \caption{w Aug. \cref{eqn:method:mppi} and $\tau=5$.}
        \label{fig:exp:overlay:aug_ua_hist5}
    \end{subfigure}
    \caption{
        \orange{Timelapse} and \cyan{trajectories} of the Spot traversing CLF using different variants of ELLIPSE.
        (a)(b): Without the domain augmentation, both variants \red{crash into handrails} due to compounding error.
        (c)(d): With the domain augmentation, both variants completes the run without help, and stay closer to stair center.
    }
    \label{fig:exp:overlay}
    \vspace{-2 em}
\end{figure} 

\begin{table}[h]
    \vspace{-1em}
    \centering
        \begin{tabular}{c | c c c c | c }
        \hline
        Model & EES & RWS & RES & CLF & Total\\
        \hline
        BEVFusion           & 4            & $\mathbf{0}$ & 1            & 3            & 8 \\
        ELLIPSE-Uni-no-Aug  & 2            & 4            & 3            & 2            & 11 \\
        ELLIPSE-Uni         & 2            & $\mathbf{0}$ & 1            & $\mathbf{0}$ & 3 \\
        ELLIPSE-no-Aug      & 3            & $\mathbf{0}$ & 2            & 3            & 8 \\
        ELLIPSE             & $\mathbf{1}$ & $\mathbf{0}$ & $\mathbf{0}$ & $\mathbf{0}$ & $\mathbf{1}$ \\
        \hline
        \end{tabular}
    \caption{
        Number of manual interventions ($\downarrow$) required to complete the 4 test sequences (\cref{fig:intro:env}). 
        Best results are shown in \textbf{bold}. 
        BEVFusion fails frequently due to inference latency. 
        ELLIPSE-Uni performs worse than ELLIPSE overall because it predicts waypoint $x$ and $y$ coordinates independently. 
        Domain augmentation substantially improves both ELLIPSE and ELLIPSE-Uni.
    }
    \label{tab:exp:sr}
    \vspace{-1.5em}
\end{table}

In this experiment, we evaluate the nominal reliability of the predicted waypoints without using uncertainty during planning.
Specifically, we interpolate the 5 predicted sparse waypoints into a dense reference path and track this path using an MPPI controller with 512 rollouts, 50 steps rollout horizon, and timestep $\Delta_t = 0.1$\,s.
The robot's linear and angular velocities are constrained to 0.5\,m/s and 1.0\,rad/s, respectively.
This tracking MPPI runs at 20\,Hz.

We compare ELLIPSE against several baselines.
BEVFusion is a widely used architecture for BEV perception~\cite{liu2022bevfusion}.
For a fairer latency comparison, we modify its architecture to use the same PointPillars backbone as our method, and train it using LiDAR input and 3 onboard RGB-D cameras (front-down, front-up, and rear).
Because synthesizing novel RGB-D observations is beyond the scope of this work, this baseline is trained without domain augmentation.
We also train a univariate deep evidential regression model~\cite{amini2020deep}, denoted ELLIPSE-Uni, and ablate domain augmentation for both the univariate and multivariate variants (denoted ELLIPSE-Uni-no-Aug and ELLIPSE-no-Aug, respectively).

\Cref{tab:exp:sr} reports the number of manual interventions required for each method to complete the four test sequences.
Although BEVFusion~\cite{liu2022bevfusion} additionally uses 3 RGB-D cameras, its performance is comparable to ELLIPSE-no-Aug.
We hypothesize that its higher inference latency ($\sim$4\,Hz) and the limited camera field of view reduced its ability to recover from out-of-distribution states.
ELLIPSE-Uni and ELLIPSE-Uni-no-Aug perform worse than their multivariate counterparts, likely because modeling the waypoint $x$ and $y$ coordinates independently fails to capture their correlation, especially during turning maneuvers on stair landings.
Most importantly, the proposed domain augmentation strategy (\cref{sec:method:aug}) significantly reduces the number of interventions for both the univariate and multivariate models.
We attribute this improvement to the fact that augmented viewpoints expose the policy to off-demonstration states and encourage corrective behavior under compounding error.
As illustrated qualitatively in \cref{fig:exp:overlay}, while ELLIPSE successfully navigates the scenario (\cref{fig:exp:overlay:aug_track}), ELLIPSE-no-Aug runs into the handrail and thus needs intervention (\cref{fig:exp:overlay:no_aug_track}).

\subsection{Empirical Coverage of the Predictive Uncertainty} \label{sec:exp:coverage}
\begin{table*}
    \centering
        \begin{tabular}{c | c | c | c | c c c c c | c c c c c }
        \hline
        Dataset & Train Aug & Calib Aug & Calib Method & \multicolumn{5}{|c|}{Empirical Coverage $\%$ ($\sim 90\%$)} & \multicolumn{5}{|c}{Sharpness $m^2$ ($\downarrow$)} \\
        \hline
        \multirow{5}{*}{Adversarial}
        & Y                  & Y & \multirow{3}{*}{Isotonic}   & 0.88             & 0.95             & \underline{0.95} & \underline{0.92} & \textbf{0.90}    & \underline{0.10} & \underline{0.29} & \underline{0.48} & \underline{0.58} & \underline{0.86} \\
        \cline{2-3}
        & \multirow{4}{*}{N} & N &                             & 0.61             & 0.70             & 0.67             & 0.64             & 0.73             & \textbf{0.06}    & \textbf{0.16}    & \textbf{0.27}    & \textbf{0.40}    & \textbf{0.72}    \\
        &                    & Y &                             & \textbf{0.90}    & \textbf{0.90}    & \textbf{0.89}    & 0.87             & \underline{0.89} & 0.22             & 0.40             & 0.60             & 0.91             & 1.57             \\
        \cline{4-14}
        &                    & N & \multirow{2}{*}{MVP}        & 0.87             & \underline{0.88} & \textbf{0.89}    & \textbf{0.90}    & 0.93             & 0.21             & 0.38             & 0.68             & 1.20             & 2.00             \\
        &                    & Y &                             & \underline{0.89} & \textbf{0.90}    & \textbf{0.89}    & 0.87             & \underline{0.89} & 0.21             & 0.39             & 0.66             & 0.94             & 1.50             \\
        \hline
        \multirow{5}{*}{Deployment}
        & Y                  & Y & \multirow{3}{*}{Isotonic}   & \underline{0.84} & \textbf{0.92}    & \textbf{0.92}    & \textbf{0.92}    & \textbf{0.90}    & \underline{0.10} & \underline{0.29} & \underline{0.49} & \underline{0.61} & \underline{0.88} \\
        \cline{2-3}
        & \multirow{4}{*}{N} & N &                             & 0.49             & 0.58             & 0.58             & 0.59             & 0.73             & \textbf{0.06}    & \textbf{0.15}    & \textbf{0.25}    & \textbf{0.38}    & \textbf{0.69}    \\
        &                    & Y &                             & 0.73             & 0.81             & 0.82             & 0.84             & \underline{0.91} & 0.22             & 0.37             & 0.58             & 0.87             & 1.52             \\
        \cline{4-14}
        &                    & N & \multirow{2}{*}{MVP}        & \textbf{0.86}    & \underline{0.86} & \underline{0.86} & \underline{0.86} & 0.88             & 0.30             & 0.54             & 0.79             & 0.94             & 1.22             \\
        &                    & Y &                             & 0.82             & 0.85             & 0.84             & 0.83             & \textbf{0.90}    & 0.27             & 0.47             & 0.67             & 0.83             & 1.41             \\
        \hline
        \end{tabular}
    \caption{
        Empirical coverage (\%) and sharpness ($\mathrm{m}^2$) of calibrated 90\% prediction sets on \textit{Adversarial} and \textit{Deployment}.
        ELLIPSE (calibrated with domain augmentation) achieves strong empirical coverage with compact prediction sets on both datasets.
        For ELLIPSE-no-Aug, calibrating on augmented data substantially improves coverage, but at the cost of much larger prediction sets.
        Although ELLIPSE-no-Aug-MVP yields coverage closest to the target 90\%, it relies on privileged online conformity feedback.
        (\textbf{Best}, and \underline{Second Best})
    }
    \label{tab:exp:cov}
    \vspace{-2em}
\end{table*}
Next, we evaluate the empirical coverage of the predicted uncertainty sets.
We split the four test staircases into a calibration set (EES and RWS) and a deployment set (RES and CLF).
Our goal is to construct calibrated 90\% prediction sets, i.e., calibrated ellipses that contain the ground-truth waypoint approximately 90\% of the time during deployment.
We additionally report sharpness, measured as the area of the resulting prediction sets~\cite{kuleshov2018accurate}.
Ideally a calibrated predictor would achieve close to 90\% empirical coverage while keeping the prediction sets as compact as possible.

As baselines, we consider ELLIPSE-no-Aug and a strong online conformal prediction baseline, Multi Valid Prediction (ELLIPSE-no-Aug-MVP)~\cite{bastani2022practical}, which provides group-wise coverage guarantees.
We use $5$ groups in MVP, i.e., a separate threshold for each waypoint index.

Since MVP is an online conformal method, it updates its thresholds during deployment using a conformity score.
Therefore we provide MVP with ground-truth waypoints online (i.e., an oracle update protocol), which favors MVP.
We evaluate ELLIPSE-no-Aug and ELLIPSE-no-Aug-MVP under two calibration regimes: calibration on clean demonstrations only, and calibration on domain-augmented demonstrations.
This comparison helps to highlight the importance of incorporating domain augmentation during calibration.

We evaluate the calibrated uncertainties on two deployment datasets.
In the first experiment, we teleoperate the robot to traverse the stairs in an adversarial manner (aggressive turning and zig-zagging; denoted \textit{Adversarial}).
In the second experiment, we deploy ELLIPSE and ELLIPSE-no-Aug on the deployment staircases and record the resulting point clouds and poses (denoted \textit{Deployment}).

The empirical coverage and sharpness of the 90\% prediction sets are reported in \cref{tab:exp:cov}.
ELLIPSE, trained and calibrated with domain augmentation, achieves high empirical coverage on both datasets.
Although it slightly over-covers the ground truth, the resulting prediction sets remain reasonably compact.
In contrast, ELLIPSE-no-Aug calibrated on clean demonstrations produces the smallest prediction sets, but severely underestimates the true residuals, resulting in the worst empirical coverage.
Calibrating ELLIPSE-no-Aug with domain-augmented data significantly improves empirical coverage, but does so by aggressively enlarging the prediction sets.
We attribute this to persistent overconfidence when the robot states are off the demonstration manifold.
Finally, although ELLIPSE-no-Aug-MVP achieves coverage closest to the 90\% target, its prediction set sizes are comparable to ELLIPSE-no-Aug, and MVP requires privileged information that is unavailable at deployment.

\subsection{Qualitative Analysis of Motion Planner} \label{sec:exp:planner}
\begin{figure}[t] 
    \centering
    \begin{subfigure}[b]{0.4\linewidth}
        \centering
        \includegraphics[width=0.95\columnwidth]{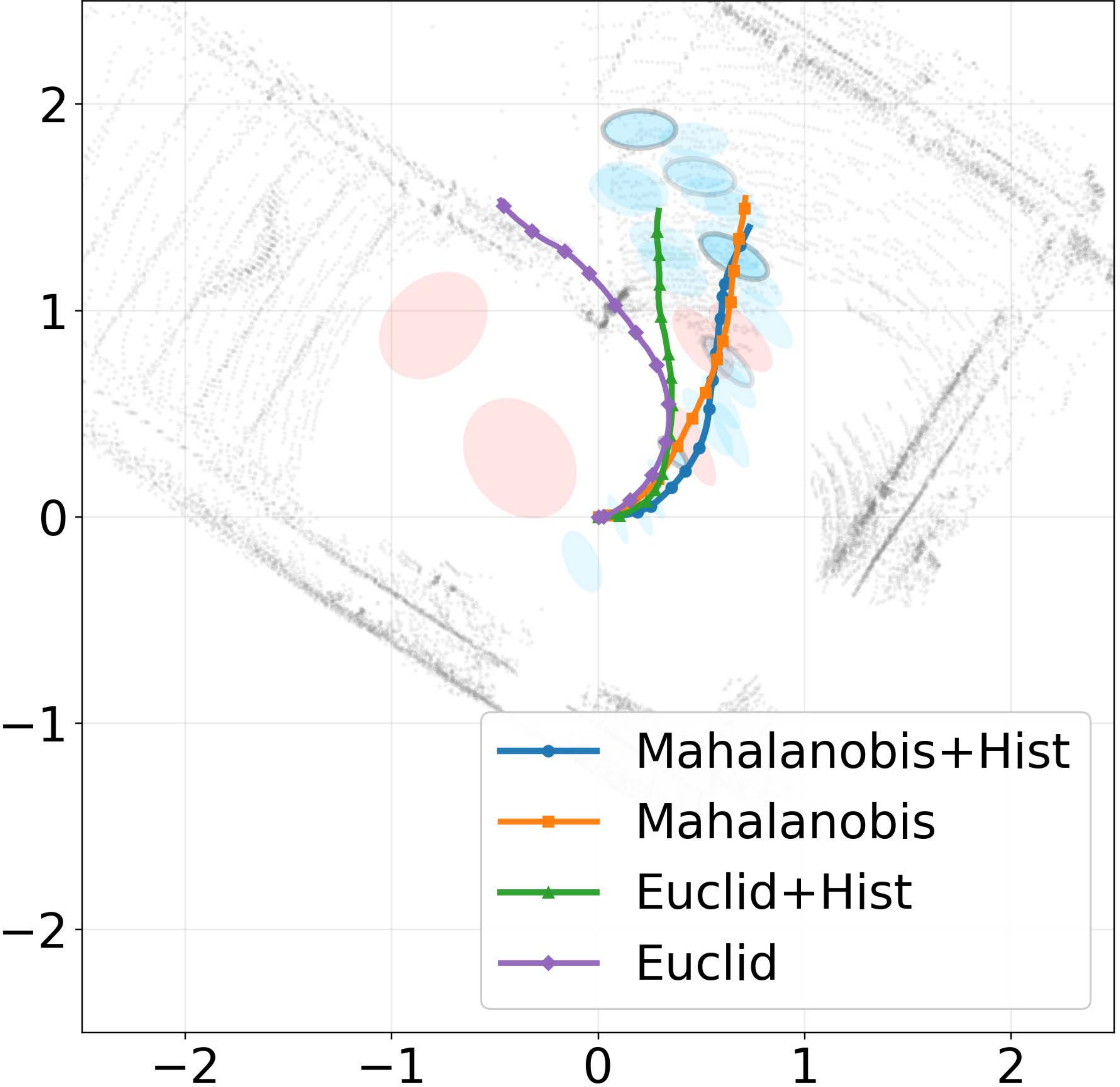}
        \vspace{-0.5em}
        \caption{}
        \label{fig:exp:mppi:left_2}
    \end{subfigure}%
    \begin{subfigure}[b]{0.4\linewidth}
        \centering
        \includegraphics[width=0.95\columnwidth]{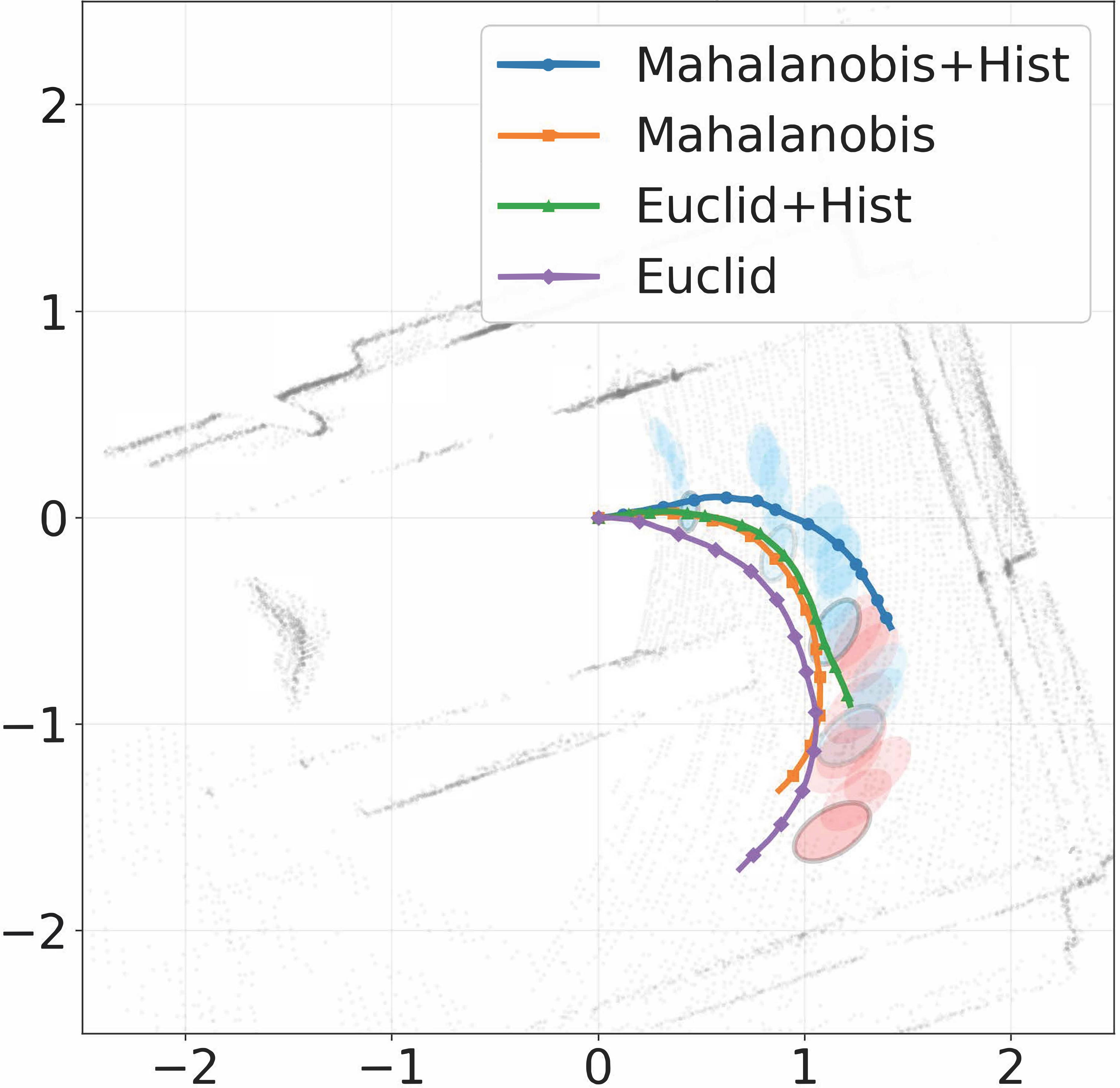}
        \vspace{-0.5em}
        \caption{}
        \label{fig:exp:mppi:right_1}
    \end{subfigure}
    \vspace{-0.5em}
    \caption{
        Qualitative comparison of MPPI planning variants, overlaid with unit-level uncertainty ellipses (\cyan{accepted}, \red{relaxed}; current-step predictions are highlighted with \textbf{black} edges).
        \textit{\blue{Mahalanobis+Hist}} remains close to confident predictions, whereas the other variants can be dominated by highly uncertain waypoints, leading to potentially unsafe behavior under disturbances.
    }
    \label{fig:exp:mppi}
    \vspace{-2em}
\end{figure} 
As discussed in \cref{sec:exp:sr}, ELLIPSE already achieves the highest success rate among the considered baselines using a simple path-tracking controller, largely due to the high control frequency of the MPPI planner (20~Hz).
However, the tracker may still deviate from the predicted waypoints due to control limits, and it is vulnerable to occasional poor predictions because it does not explicitly use waypoint uncertainty or historical confident predictions.
We therefore evaluate the uncertainty-aware MPPI planner in \cref{sec:method:mppi}. 
Since all variants achieve near-saturated success rates, we focus on qualitative comparisons.
Specifically, we compare four planner variants with different cost formulations:
\begin{itemize}
    \item \textit{\blue{Mahalanobis+Hist}}: \cref{eqn:method:mppi} with history ($\tau=5$),
    \item \textit{\orange{Mahalanobis}}: \cref{eqn:method:mppi} without history ($\tau=0$),
    \item \textit{\green{Euclid+Hist}}: Minimize Euclid. Dist to waypoints with history ($\tau=5$),
    \item \textit{\purple{Euclid}}: Minimize Euclid. Dist to waypoints without history (the best path-tracking baseline in \cref{sec:exp:sr}).
\end{itemize}
For \textit{\blue{Mahalanobis+Hist}} and \textit{\orange{Mahalanobis}}, we use $\delta=0.2$~m and $\beta=2.0$.
Example results are shown in \cref{fig:exp:mppi}.

Because \textit{\purple{Euclid}} and \textit{\green{Euclid+Hist}} weight all waypoints equally, they may converge to aggressively turning paths that pass too close to obstacles (e.g., stair handrails; \cref{fig:exp:mppi:left_2,fig:exp:mppi:right_1}).
Although \textit{\orange{Mahalanobis}} can relax uncertain predictions, it only uses the current prediction and may still fail when multiple current waypoints are uncertain (\cref{fig:exp:mppi:right_1}).
In contrast, \textit{\blue{Mahalanobis+Hist}} stays close to confident waypoints and behaves conservatively when most predictions are uncertain, improving robustness to occasional errors and disturbances.
Another qualitative example is visualized in (\cref{fig:exp:overlay}), where \textit{\blue{Mahalanobis+Hist}} keeps the robot closer to staircase center (\cref{fig:exp:overlay:aug_ua_hist5}) compared to \textit{\purple{Euclid}} (\cref{fig:exp:overlay:aug_track}).
On the other hand, even with \textit{\blue{Mahalanobis+Hist}}, ELLIPSE-no-Aug still runs into obstacles because it assigns high confidence to wrong predictions (\cref{fig:exp:overlay:no_aug_track_hist5}).

\section{Conclusion}

In this paper, we present ELLIPSE, a system that predicts uncertainty-aware waypoints building on multivariate deep evidential regression~\cite{meinert2021multivariate}.
To improve the robustness of the waypoints and uncertainties, we employ a simple point cloud-based domain augmentation that synthesizes new instances from viewpoints away from the demonstration trajectories. 
The uncertainties from the model is recalibrated with an isotonic regression module to adjust them to deployment error magnitudes.
The waypoints and uncertainties are integrated into an uncertainty-aware MPPI planner that relaxes tracking around highly uncertain waypoints.
Through extensive quantitative and qualitative experiments, we demonstrate the practical benefits of ELLIPSE over several baselines.
Limitations and future works include evaluating on more tasks, and integration with online label-free calibration.



\printbibliography

@article{nahavandi2025comprehensive,
  title={A comprehensive review on autonomous navigation},
  author={Nahavandi, Saeid and Alizadehsani, Roohallah and Nahavandi, Darius and Mohamed, Shady and Mohajer, Navid and Rokonuzzaman, Mohammad and Hossain, Ibrahim},
  journal={ACM Computing Surveys},
  volume={57},
  number={9},
  pages={1--67},
  year={2025},
  publisher={ACM New York, NY}
}

@inproceedings{sridhar2024nomad,
  title={Nomad: Goal masked diffusion policies for navigation and exploration},
  author={Sridhar, Ajay and Shah, Dhruv and Glossop, Catherine and Levine, Sergey},
  booktitle={2024 IEEE International Conference on Robotics and Automation (ICRA)},
  pages={63--70},
  year={2024},
  organization={IEEE}
}

@article{yu2024trajectory,
  title={Trajectory diffusion for objectgoal navigation},
  author={Yu, Xinyao and Zhang, Sixian and Song, Xinhang and Qin, Xiaorong and Jiang, Shuqiang},
  journal={Advances in Neural Information Processing Systems},
  volume={37},
  pages={110388--110411},
  year={2024}
}

@inproceedings{koller2018learning,
  title={Learning-based model predictive control for safe exploration},
  author={Koller, Torsten and Berkenkamp, Felix and Turchetta, Matteo and Krause, Andreas},
  booktitle={2018 IEEE conference on decision and control (CDC)},
  pages={6059--6066},
  year={2018},
  organization={IEEE}
}

@article{dong2024collision,
  title={Collision avoidance verification of multiagent systems with learned policies},
  author={Dong, Zihao and Omidshafiei, Shayegan and Everett, Michael},
  journal={IEEE Control Systems Letters},
  volume={8},
  pages={652--657},
  year={2024},
  publisher={IEEE}
}

@article{dong2025learning,
  title={Learning Smooth State-Dependent Traversability from Dense Point Clouds},
  author={Dong, Zihao and Papalia, Alan and Jung, Leonard and Spiro, Alenna and Osteen, Philip R and Robison, Christa S and Everett, Michael},
  journal={arXiv preprint arXiv:2506.04362},
  year={2025}
}

@article{abdar2021review,
  title={A review of uncertainty quantification in deep learning: Techniques, applications and challenges},
  author={Abdar, Moloud and Pourpanah, Farhad and Hussain, Sadiq and Rezazadegan, Dana and Liu, Li and Ghavamzadeh, Mohammad and Fieguth, Paul and Cao, Xiaochun and Khosravi, Abbas and Acharya, U Rajendra and others},
  journal={Information fusion},
  volume={76},
  pages={243--297},
  year={2021},
  publisher={Elsevier}
}

@article{he2025survey,
  title={A survey on uncertainty quantification methods for deep learning},
  author={He, Wenchong and Jiang, Zhe and Xiao, Tingsong and Xu, Zelin and Li, Yukun},
  journal={ACM Computing Surveys},
  year={2025},
  publisher={ACM New York, NY}
}

@article{cai2024evora,
  title={Evora: Deep evidential traversability learning for risk-aware off-road autonomy},
  author={Cai, Xiaoyi and Ancha, Siddharth and Sharma, Lakshay and Osteen, Philip R and Bucher, Bernadette and Phillips, Stephen and Wang, Jiuguang and Everett, Michael and Roy, Nicholas and How, Jonathan P},
  journal={IEEE Transactions on Robotics},
  year={2024},
  publisher={IEEE}
}

@article{zhao2024conformalized,
  title={Conformalized interactive imitation learning: Handling expert shift and intermittent feedback},
  author={Zhao, Michelle and Simmons, Reid and Admoni, Henny and Ramdas, Aaditya and Bajcsy, Andrea},
  journal={arXiv preprint arXiv:2410.08852},
  year={2024}
}

@article{bastani2022practical,
  title={Practical adversarial multivalid conformal prediction},
  author={Bastani, Osbert and Gupta, Varun and Jung, Christopher and Noarov, Georgy and Ramalingam, Ramya and Roth, Aaron},
  journal={Advances in neural information processing systems},
  volume={35},
  pages={29362--29373},
  year={2022}
}

@article{angelopoulos2021gentle,
  title={A gentle introduction to conformal prediction and distribution-free uncertainty quantification},
  author={Angelopoulos, Anastasios N and Bates, Stephen},
  journal={arXiv preprint arXiv:2107.07511},
  year={2021}
}

@article{tibshirani2019conformal,
  title={Conformal prediction under covariate shift},
  author={Tibshirani, Ryan J and Foygel Barber, Rina and Candes, Emmanuel and Ramdas, Aaditya},
  journal={Advances in neural information processing systems},
  volume={32},
  year={2019}
}

@article{ovadia2019can,
  title={Can you trust your model's uncertainty? evaluating predictive uncertainty under dataset shift},
  author={Ovadia, Yaniv and Fertig, Emily and Ren, Jie and Nado, Zachary and Sculley, David and Nowozin, Sebastian and Dillon, Joshua and Lakshminarayanan, Balaji and Snoek, Jasper},
  journal={Advances in neural information processing systems},
  volume={32},
  year={2019}
}

@article{amini2020deep,
  title={Deep evidential regression},
  author={Amini, Alexander and Schwarting, Wilko and Soleimany, Ava and Rus, Daniela},
  journal={Advances in neural information processing systems},
  volume={33},
  pages={14927--14937},
  year={2020}
}

@article{meinert2021multivariate,
  title={Multivariate deep evidential regression},
  author={Meinert, Nis and Lavin, Alexander},
  journal={arXiv preprint arXiv:2104.06135},
  year={2021}
}

@inproceedings{ryu2023instant,
  title={Instant domain augmentation for lidar semantic segmentation},
  author={Ryu, Kwonyoung and Hwang, Soonmin and Park, Jaesik},
  booktitle={Proceedings of the IEEE/CVF Conference on Computer Vision and Pattern Recognition},
  pages={9350--9360},
  year={2023}
}

@inproceedings{bramlage2023plausible,
  title={Plausible uncertainties for human pose regression},
  author={Bramlage, Lennart and Karg, Michelle and Curio, Crist{\'o}bal},
  booktitle={Proceedings of the IEEE/CVF International Conference on Computer Vision},
  pages={15133--15142},
  year={2023}
}

@article{hu2024orbitgrasp,
  title={OrbitGrasp: $ SE (3) $-Equivariant Grasp Learning},
  author={Hu, Boce and Zhu, Xupeng and Wang, Dian and Dong, Zihao and Huang, Haojie and Wang, Chenghao and Walters, Robin and Platt, Robert},
  journal={arXiv preprint arXiv:2407.03531},
  year={2024}
}

@article{cheng2024navila,
  title={Navila: Legged robot vision-language-action model for navigation},
  author={Cheng, An-Chieh and Ji, Yandong and Yang, Zhaojing and Gongye, Zaitian and Zou, Xueyan and Kautz, Jan and B{\i}y{\i}k, Erdem and Yin, Hongxu and Liu, Sifei and Wang, Xiaolong},
  journal={arXiv preprint arXiv:2412.04453},
  year={2024}
}

@inproceedings{ross2011reduction,
  title={A reduction of imitation learning and structured prediction to no-regret online learning},
  author={Ross, St{\'e}phane and Gordon, Geoffrey and Bagnell, Drew},
  booktitle={Proceedings of the fourteenth international conference on artificial intelligence and statistics},
  pages={627--635},
  year={2011},
  organization={JMLR Workshop and Conference Proceedings}
}

@inproceedings{menda2019ensembledagger,
  title={Ensembledagger: A bayesian approach to safe imitation learning},
  author={Menda, Kunal and Driggs-Campbell, Katherine and Kochenderfer, Mykel J},
  booktitle={2019 IEEE/RSJ International Conference on Intelligent Robots and Systems (IROS)},
  pages={5041--5048},
  year={2019},
  organization={IEEE}
}

@book{neal2012bayesian,
  title={Bayesian learning for neural networks},
  author={Neal, Radford M},
  volume={118},
  year={2012},
  publisher={Springer Science \& Business Media}
}

@book{zhou2025ensemble,
  title={Ensemble methods: foundations and algorithms},
  author={Zhou, Zhi-Hua},
  year={2025},
  publisher={Chapman and Hall/CRC}
}

@article{gibbs2024conformal,
  title={Conformal inference for online prediction with arbitrary distribution shifts},
  author={Gibbs, Isaac and Cand{\`e}s, Emmanuel J},
  journal={Journal of Machine Learning Research},
  volume={25},
  number={162},
  pages={1--36},
  year={2024}
}

@inproceedings{meinert2023unreasonable,
  title={The unreasonable effectiveness of deep evidential regression},
  author={Meinert, Nis and Gawlikowski, Jakob and Lavin, Alexander},
  booktitle={Proceedings of the AAAI Conference on Artificial Intelligence},
  volume={37},
  number={8},
  pages={9134--9142},
  year={2023}
}

@inproceedings{kuleshov2018accurate,
  title={Accurate uncertainties for deep learning using calibrated regression},
  author={Kuleshov, Volodymyr and Fenner, Nathan and Ermon, Stefano},
  booktitle={International conference on machine learning},
  pages={2796--2804},
  year={2018},
  organization={PMLR}
}

@article{mandlekar2023mimicgen,
  title={Mimicgen: A data generation system for scalable robot learning using human demonstrations},
  author={Mandlekar, Ajay and Nasiriany, Soroush and Wen, Bowen and Akinola, Iretiayo and Narang, Yashraj and Fan, Linxi and Zhu, Yuke and Fox, Dieter},
  journal={arXiv preprint arXiv:2310.17596},
  year={2023}
}

@inproceedings{zhou2023nerf,
  title={Nerf in the palm of your hand: Corrective augmentation for robotics via novel-view synthesis},
  author={Zhou, Allan and Kim, Moo Jin and Wang, Lirui and Florence, Pete and Finn, Chelsea},
  booktitle={Proceedings of the IEEE/CVF Conference on Computer Vision and Pattern Recognition},
  pages={17907--17917},
  year={2023}
}

@article{oh2025self,
  title={Self-Augmented Robot Trajectory: Efficient Imitation Learning via Safe Self-augmentation with Demonstrator-annotated Precision},
  author={Oh, Hanbit and Murooka, Masaki and Motoda, Tomohiro and Nakajo, Ryoichi and Domae, Yukiyasu},
  journal={arXiv preprint arXiv:2509.09893},
  year={2025}
}

@article{chen2022direct,
  title={Direct lidar-inertial odometry: Lightweight lio with continuous-time motion correction},
  author={Chen, Kenny and Nemiroff, Ryan and Lopez, Brett T},
  journal={arXiv preprint arXiv:2203.03749},
  year={2022}
}

@inproceedings{dong2025lidar,
  title={Lidar inertial odometry and mapping using learned registration-relevant features},
  author={Dong, Zihao and Pflueger, Jeff and Jung, Leonard and Thorne, David and Osteen, Philip R and Robison, Christa S and Lopez, Brett T and Everett, Michael},
  booktitle={2025 IEEE International Conference on Robotics and Automation (ICRA)},
  pages={359--366},
  year={2025},
  organization={IEEE}
}

@article{belkhale2023data,
  title={Data quality in imitation learning},
  author={Belkhale, Suneel and Cui, Yuchen and Sadigh, Dorsa},
  journal={Advances in neural information processing systems},
  volume={36},
  pages={80375--80395},
  year={2023}
}

@book{roth2012multivariate,
  title={On the multivariate t distribution},
  author={Roth, Michael},
  year={2012},
  publisher={Link{\"o}ping University Electronic Press}
}

@inproceedings{williams2016aggressive,
  title={Aggressive driving with model predictive path integral control},
  author={Williams, Grady and Drews, Paul and Goldfain, Brian and Rehg, James M and Theodorou, Evangelos A},
  booktitle={2016 IEEE international conference on robotics and automation (ICRA)},
  pages={1433--1440},
  year={2016},
  organization={IEEE}
}

@inproceedings{lang2019pointpillars,
  title={Pointpillars: Fast encoders for object detection from point clouds},
  author={Lang, Alex H and Vora, Sourabh and Caesar, Holger and Zhou, Lubing and Yang, Jiong and Beijbom, Oscar},
  booktitle={Proceedings of the IEEE/CVF conference on computer vision and pattern recognition},
  pages={12697--12705},
  year={2019}
}

@article{liu2022bevfusion,
  title={Bevfusion: Multi-task multi-sensor fusion with unified bird's-eye view representation},
  author={Liu, Zhijian and Tang, Haotian and Amini, Alexander and Yang, Xinyu and Mao, Huizi and Rus, Daniela and Han, Song},
  journal={arXiv preprint arXiv:2205.13542},
  year={2022}
}

\end{document}